\documentclass[10pt,letterpaper]{article}

\usepackage{cogsci}
\usepackage{booktabs}
\usepackage{enumitem}
\usepackage[title]{appendix}
\usepackage{multicol}
\usepackage{tabularx}
\usepackage{graphicx}
\usepackage{hyperref}

\cogscifinalcopy %%% Uncomment this line for the final submission

\usepackage[
  style=apa,
  natbib=true,
  annotation=false,
]{biblatex}
\usepackage{float} %%% Roger Levy added this and changed figure/table placement to [H] for conformity to Word template, though floating tables and figures to top is still generally recommended!

\title{Computational Phenomenology of Temporal Experience in Autism: Quantifying the Emotional and Narrative Characteristics of Lived Unpredictability}

\author[1]{\mbox{Kacper Dudzic (kacper.dudzic@ideas.edu.pl)}}
\author[2]{\mbox{Karolina Drożdż}}
\author[3]{\mbox{Maciej Wodziński}}
\author[4]{\mbox{Anastazja Szuła}}
\author[5]{\mbox{Marcin Moskalewicz}}
\affil[1]{Adam Mickiewicz University}
\affil[1]{AMU Center for Artificial Intelligence}
\affil[1,2,5]{IDEAS Research Institute}
\affil[3]{Maria Curie-Skłodowska University}
\affil[4,5]{Poznań University of Medical Sciences}

\begin{document}
\raggedbottom

\maketitle

\begin{abstract}
Disturbances in temporality, such as desynchronization with the social environment and its unpredictability, are considered core features of autism with a deep impact on relationships. However, limitations regarding research on this issue include: 1) the dominance of deficit-based medical models of autism, 2) sample size in qualitative research, and 3) the lack of phenomenological anchoring in computational research. To bridge the gap between phenomenological and computational approaches and overcome sample-size limitations, our research integrated three methodologies. Study A: structured phenomenological interviews with autistic individuals using the Transdiagnostic Assessment of Temporal Experience ($N = 28$). Study B: computational analysis of an autobiographical corpus of autistic narratives (ca. 7 million words from 104 authors) built for this purpose. We developed a lexicon of 80 temporal adverbs and employed aspect-based sentiment analysis to compare their occurrences in the corpus against general time expressions. Study C: a replication of a computational study using narrative flow measures to assess the perceived phenomenological authenticity of autistic autobiographies. Interviews revealed that the most significant differences between the autistic and control groups concerned unpredictability of experience. Computational results mirrored these findings: the temporal lexicon in autistic narratives was significantly more negatively valenced---particularly the ,,Immediacy \& Suddenness'' category. Outlier analysis identified terms associated with perceived discontinuity (\textit{unpredictably}, \textit{precipitously}, and \textit{abruptly}) as highly negative. The computational analysis of narrative flow found that the autistic narratives contained within the corpus quantifiably resemble autobiographical stories more than imaginary ones. Overall, the temporal challenges experienced by autistic individuals were shown to primarily concern lived unpredictability and stem from the contents of lived experience, and not from autistic narrative construction.

\end{abstract}

\section{Introduction}

\textbf{Autism: a disorder or condition?} Diagnostic medical frameworks define Autism Spectrum Disorder (ASD) through observable difficulties in communication and social interaction, alongside restricted or repetitive behavioral patterns \parencite{APA2013, WHO2004}. At the same time, public discourse on ASD is predominantly constructed through deficit-based narratives that prioritize neurotypical and medical perspectives over lived experience. The widespread narrative depicts autism through a ,,bio-medical'' or ,,socio-negative'' lens \parencite{huws2011, Wodzinski2025, Wolbring2017}, focusing on ,,abnormalities'' and the necessity of correction, while the public perpetuates the negative associations despite accurate knowledge \parencite{Billawala2014, Wodzinski2021}. Consequently, the public image of autism remains biased, as media representations often exclude autistic voices in favor of neurotypical journalists or health professionals, who further perpetuate stereotypical views. Additionally, the medical approach, relying solely on behavioral markers, fails to fully capture the complex, subjective reality of how autistic individuals perceive the world and relate to others \parencite{Fuchs2015, Fuchs2018}. While the terms \textit{disorder} and \textit{condition} are often used interchangeably in the literature, this paper adopts the concept of Autism Spectrum Condition (ASC) to avoid the deficit-based implications of the psychiatric medical model and align with the neurodiversity movement's approach (which frames autism not as a mental health issue to be fixed, but as an alternative mode of experience) \parencite{BaronCohen2015}. Our approach is also motivated by the fact that the paper was co-written by people with autism, the so-called ,,experts by experience.'' 

\textbf{Experience of time in autism.} Temporal disturbances are frequently seen as the core feature of many psychopathological phenomena \parencite{Botbol2013, Fuchs2001, Fuchs2015, Molder2016, Moskalewicz2016, Nielsen2017, Vogel2020, Wyllie2005}. In this respect, autism is characterized mainly by desynchronization with the social world, the perception of others' actions as unpredictable, difficulties with planning, repetitive schemas, and difficulty seeing cause-and-effect relationships \parencite{Vogel2020, Korendo2024}. Some researchers suggest that autism can be re-conceptualized through the lens of temporality \parencite{Christensen2021, Green2022, Lopez2015, Boldsen2018, Boldsen2022}; this approach would inform integrative research across quantitative and qualitative methodologies \parencite{Orm2021}, understanding how these temporal frames shape decision-making \parencite{Fujino2020}, emotional well-being \parencite{AlHatmi2015}, and identity \parencite{Cohen2022, Xuan2022}. Notably, routine and predictability are often reported to provide autistic individuals with a sense of temporal structure. Predictability as an ability refers to the degree to which an individual can foresee upcoming events, the behavior of others, and the consequences of their own actions. This ability facilitates the integration of past experiences with the present context to inform future actions or understanding of social cues \parencite{Vogel2019} and can be disrupted by, e.g., overwhelming sensory experiences \parencite{Taels2023}. First-person accounts of autism highlight difficulties in ,,predicting the world'', a sense of being ,,stuck in the present'', and reliance on routine as a way to manage an uncertain flow of time \parencite{Boucher2001, Brunetti2022, Coelho2022, Poole2024, Spikins2016, Zukauskas2001, Zukauskas2009}. Studies also indicate that the present experience in autism is often strongly shaped by past experiences with limited spontaneity \parencite{Zukauskas2001}. All in all, the mismatch with cultural norms of rhythmicity can result in feeling out of sync with societal expectations, even when there is no real-time pressure \parencite{Jensen2017}. For example, autistic children often experience the world as moving too fast and struggle with fragmented autobiographical memory, which disrupts their sense of temporal continuity and affects the development of identity, bodily self, and social-emotional relationships \parencite{Pelloux2024, Feller2023}. Their personal rhythms do not align with conventional schedules \parencite{Christensen2021}.

\textbf{The role of phenomenology.} Although disturbances of temporal experience are central to phenomenological accounts of many psychiatric conditions---including autism---there remains a gap between philosophical descriptions of the ,,essence'' of atypical experiences and empirical evidence on their incidence and severity as assessed with validated psychological and clinical measures. Phenomenological evidence is frequently speculative, anecdotal, and, when qualitative and empirical, it is typically based on small samples. Nevertheless, such evidence has the advantage of being close to lived subjective experience, which is typically gathered through direct first-person reports, and not objectifying surveys. Research on media representations of autism shows that utilizing such a subjective perspective is a methodological necessity for constructing a non-stigmatizing understanding of the autistic experience. First-person testimonies, despite being marginally represented (appearing only in 3.7\% of media reports), significantly shift the discursive tone from negative to both positive and more nuanced \parencite{Wodzinski2025}. This stands in stark contrast to the contributions of health professionals, who reinforce a socio-negative, deficit-based image. A key challenge is how to scale this approach when qualitative samples remain small while quantitative methods depart from subjective experience.

One solution targeting the validity of phenomenological findings is the Transdiagnostic Assessment of Temporal Experience (TATE), a recently developed structured phenomenological interview designed to systematically assess multiple aspects of lived time in a quantitative format \parencite{Stanghellini2022}. Another solution, targeting the limitations of qualitative samples, is a large corpus of first-person descriptions of autistic experiences, which could be used as a textual resource for computational research on ASC subjective data. We have built such a corpus consisting entirely of selected autobiographical writings by autistic authors---the ASC Autobiographical Corpus\footnote{\url{https://github.com/kdudzic/asc-autobiographical-corpus-public}}. It comprises 104 autobiographies published between 1986 and 2023, totaling nearly 7 million words (for basic statistics, see Table~\ref{tab:corpus_stats}). The collection represents a broad demographic spectrum across age, ethnicity, and gender. Notably, 58\% of the sources are authored or co-authored by female and non-binary individuals---groups historically marginalized within the broader ASC community.

\textbf{Aims of research.} The overall aim of this study is to bridge the gap between phenomenological and computational methodologies in research on subjective autistic experience. Study A is based on TATE and concerns $N = 28$ ASC individuals assessed against a control group regarding the severity of their temporal experience. Study B, a follow-up to TATE results, focuses on the sentiment of temporal semantics in autism using our autobiographical corpus. Study C adds another layer to the previous findings through a replication of research by \textcite{sap2022} on sequentiality---a computational measure of narrative flow---on the same corpus to measure whether autistic narratives’ flow resembles that of neurotypicals.

\begin{table}[h]
    \centering
    \begin{tabular}{lrrr}
        \hline
        Text Unit  & Average & Median & Total   \\
        \hline
        Sentences  & 3664.88              & 3221.50             & 381148  \\
        Words      & 65685.15             & 62102               & 6831256 \\
        Characters & 367930.59            & 349687              & 38264781\\
        \hline
    \end{tabular}
    \caption{Summary statistics of text units in the ASC Autobiographical Corpus.}
    \label{tab:corpus_stats}
\end{table}

\section{Experimental Studies}
\subsection{Study A: Methods}
\textbf{Transdiagnostic Assessment of Temporal Experience (TATE)}. We used a Polish version of TATE, recently adapted and validated \parencite{Szula2024}. TATE comprises 42 items grouped into seven categories reflecting key dimensions of lived temporality: anomalies of synchrony (1.a--1.c), time structure (2.a--2.c), implicit time flow (3.a--3.f), explicit time flow (4.a--4.c), and anomalous experiences of the past (5.a--5.f), present (6.a--6.b), and future (7.a--7.g). Participants rate each item on three 0--7 scales assessing frequency, intensity, and impairment. A severity score is calculated by averaging these three ratings, and the total TATE score (range 0--294) is obtained by summing all item severity scores.

\textbf{Study group.} A total of $N = 28$ participants were recruited. Inclusion required a clinical autism diagnosis supported by ADOS-2 or ADI-R, and the study excluded individuals with comorbid ADHD. Control data came from the Polish TATE validation study \parencite{Szula2024}, which included 98 students from the Poznan University of Medical Sciences without diagnosed mental disorders (78 female, 20 male; age 18--31, $M = 20.7, SD = 2.0$). Data were collected through interviews, each lasting 30–60 minutes. The interviewer read each prompt aloud and displayed it on screen together with the three rating scales. Participants discussed each item and selected their responses, which remained visible throughout the interview.

\subsection*{Study A: Results}
\label{subsec:study_a_results}

A principal component analysis showed a one-factor solution across domains, accounting for 29.26\% of the total variance. Items 3.f, 4.a, 4.c, 5.b, 6.h, 6.e, 7.e, 7.d, and 7.f loaded on this factor, which reflects the structure of ASC temporality. In addition, group differences were tested using the Mann–Whitney U test. The largest differences between the study and control groups were found for items 5.f (,,Reviving the past''; $p < .001, \eta^2 = .15$) and 6.k (,,Falling from the sky''; $p < .001, \eta^2 = .20$) (see Appendix~\ref{app:factor_analysis}). Item 5.f describes \textit{the past entering the present with immediacy}, while item 6.k refers to \textit{sudden discontinuities of presence}. These findings are consistent with existing research, as both items capture immediacy and suddenness of experience and thus pertain to the unpredictability of temporal experience in autism.
 
\subsection*{Study B: Methods}

\paragraph{ASC Autobiographical Corpus.} To construct the corpus, we first manually identified and subsequently acquired PDF files of existing ASC biographies, followed by the creation of a separate metadata file containing all relevant bibliographic information about the sources. Next, we manually inspected the acquired sources and marked the page ranges containing each source's main textual content in the metadata file. A co-author of the study with lived experience of autism subsequently manually cross-referenced the PDF files' contents with the metadata, filtering out any titles deemed not to be autobiographical and making corrections to the page ranges where necessary. Finally, a Python script was used to de-duplicate the collected titles, extract the text from within the page ranges from the PDF files (using the PyMuPDF library), and finally clean the extracted text by removing page numbers, mid-word line breaks, as well as extra whitespace characters.

\textbf{Temporal Lexicon.} Based on the results of TATE in \hyperref[subsec:study_a_results]{Study A}, we developed a specialized lexicon to capture how time is perceived and navigated. The selection of words was directly guided by two qualitative prompts from TATE, namely items 5.f and 6.k (see Appendix~\ref{app:tate_items}) \parencite{Szula2024}, and in this sense operationalized these items. This inventory was constructed by synthesizing data from multiple sources: an analysis of time-perception narratives in both neurodiverse and neurotypical populations, keyword searches across standard lexicographical resources, and expert consensus, resulting in 80 adverbs that represent temporal experience. Subsequently, and now independently of TATE, these terms were organized into six distinct semantic categories: Immediacy \& Suddenness, Frequency \& Repetition, Duration \& Time Blindness, Sequence \& Relative Time, Pace, and Overlap (see Appendix~\ref{app:lexicon}). We used spaCy’s \parencite{honnibal2020spacy} \texttt{EntityRuler} component added to its \texttt{en\_core\_web\_trf} English text processing pipeline to integrate all the adverbs from the lexicon into the pipeline’s entity dictionary, and then identified and marked their total of 97,192 occurrences in the pipeline-processed version of the corpus.

\textbf{General Time Expression Selection.} To establish a comparative baseline, we constructed a control group of ,,general time expressions'' through a data-driven approach. Using the same spaCy pipeline's default NER component, we gathered and marked all entities assigned the \texttt{DATE} and \texttt{TIME} labels from the pipeline-processed corpus. From this initial list, we removed any adverbs that overlapped with our specialized lexicon to ensure the two groups were mutually exclusive. Second, we filtered out idiosyncratic or rare usage. Specifically, we applied an a priori frequency cut-off, retaining only those general time expressions with a minimum appearance count of 50. This process yielded a final control group of $N = 172$ expressions with a total of 49,879 occurrences.

\textbf{Sentiment Polarity Calculation.} For each identified temporal expression from both groups, we extracted a contextual sentence triplet, which consisted of the sentence containing, preceding, and following the expression. If triplet creation was impossible (e.g., the very beginning or end of a book), a sentence pair was created instead. For multiple temporal expressions, we extracted the same sentence triplet multiple times, marking a different expression in each. We used the \texttt{deberta-v3-base-absa-v1.1}\footnote{\url{https://huggingface.co/yangheng/deberta-v3-base-absa-v1.1}} aspect-based sentiment analysis model to process each of the triplets, calculating a sentiment score for three polarities (negative, neutral, and positive) with respect to the temporal expression marked therein as the aspect. Four adverbs comprising the lexicon group (\textit{cyclically}, \textit{recurrently}, \textit{transiently}, and \textit{unforeseeably}) did not appear in the ASC corpus, and as such were excluded from further analysis. We then compared the sentiment profiles (negative, positive, and neutral) of the temporal lexicon with those of the control group of general time expressions. Due to the unequal sample sizes between the two groups and the unequal variances in their sentiment distributions, we employed Welch’s t-test.

\subsection*{Study B: Results}

\paragraph{Temporal Semantics in the ASC Corpus.} Negative sentiments across the temporal lexicon ($N = 76$) were significantly higher ($M = 0.44, SD = 0.13$, min-max 0.20--0.86, $IQR = 0.14$) than the positive sentiments ($M = 0.26, SD = 0.12$, min-max 0.01--0.57, $IQR = 0.13$), $t(75) = 8.82, p < .001$). Outlier screening using the IQR method identified three extremely high-value outliers for negative sentiment: \textit{precipitously} (0.86), \textit{unpredictably} (0.83), and \textit{abruptly} (0.83), representing approximately 3.9\% of the sample. The same method identified three high-value positive sentiment outliers: \textit{swiftly} (0.57), \textit{instantaneously} (0.57), and \textit{instantly} (0.56). Appendix~\ref{app:sentiment_dist} Figure~\ref{fig:sentiment_boxplot} highlights the distribution difference between the two sentiments.

It is noteworthy that all three negative outliers relate to the experience of unpredictability. It is equally noteworthy that all three positive outliers concern the lack thereof. \textit{Precipitous}, \textit{unpredictable}, and \textit{abrupt} qualitative contents refer to perceived discontinuity, whereas \textit{swift}, \textit{instantaneous}, and \textit{instant} refer to experiences that do come about smoothly (for examples from the corpus, see Appendix~\ref{app:examples}, Table~\ref{tab:examples}).

\textbf{Lexical Distribution and Frequency Across Temporal Categories.} The analysis of the temporal inventory ($N = 76$ unique adverbs) reveals significant variation in both lexical diversity and usage frequency across the six temporal categories (see Appendix~\ref{app:frequency}, Figure~\ref{fig:frequency}). The category of Sequence \& Relative Time is the most prominent, accounting for 49,504 of total mentions and the greatest lexical diversity (23 unique adverbs). This is followed by Frequency \& Repetition, which comprises 19 unique adverbs and 33,818 total mentions. The Duration \& Time Blindness category contributed 8,355 mentions across 15 unique adverbs. The categories of Pace (2 unique adverbs) and Immediacy \& Suddenness (12 unique adverbs) demonstrate comparatively low usage frequencies (2,460 and 2,915 mentions, respectively) relative to the dominant categories. The Overlap category represents the least mentioned dimension, containing only two unique adverbs and the lowest frequency of occurrence (140 mentions).

\begin{figure}[htbp]
    \centering
    % strictly limit content to the width of the single column
    \includegraphics[width=\columnwidth, keepaspectratio]{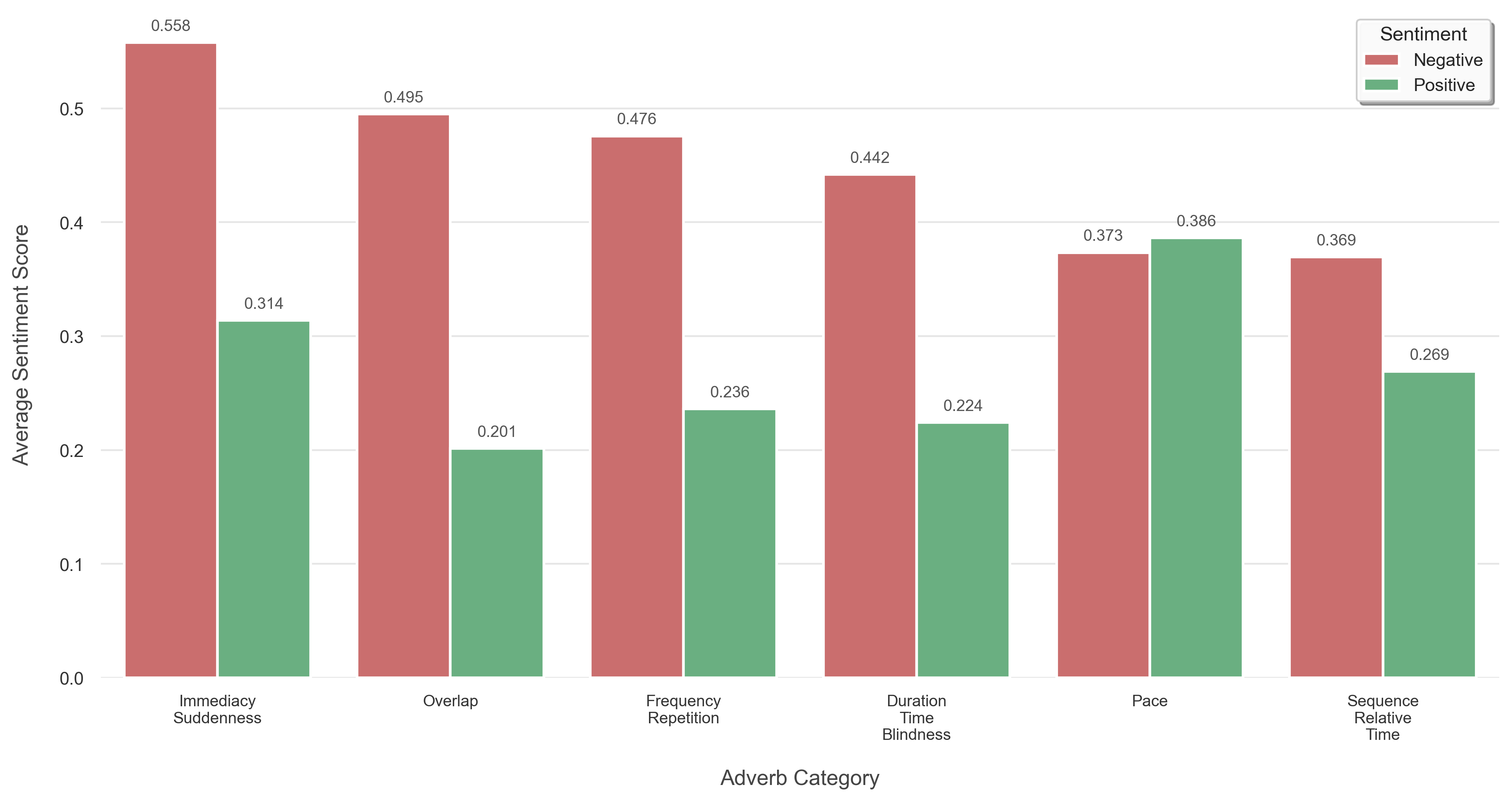}
    \caption{Distribution of positive and negative sentiment across temporal categories.}
    \label{fig:sentiment}
\end{figure}

\textbf{Sentiment Analysis of Temporal Categories.} Negative sentiment dominates across most temporal categories (see Figure~\ref{fig:sentiment}). The Immediacy \& Suddenness category has the highest average negative score (0.56), much higher than its positive score (0.31). A similar pattern is observed for other categories, i.e., Overlap (Negative: 0.5 vs. Positive: 0.2), Frequency \& Repetition (Negative: 0.48 vs. Positive: 0.24), and Duration \& Time Blindness (Negative: 0.44 vs. Positive: 0.22). While Sequence \& Relative Time is the most frequently used category, it still shows an overall negative polarity (Negative: 0.37 vs. Positive: 0.27), though this imbalance is less pronounced than in other categories. Pace stands as the only exception, displaying a rather balanced sentiment profile, with the average positive sentiment (0.39) marginally exceeding the negative (0.37).

\textbf{Comparative Sentiment Analysis.} To determine if our temporal lexicon reflects a unique affective profile, we compared its sentiment scores ($N = 76$) against the control group of general time expressions ($N = 172$) using Welch’s t-test. The analysis revealed that the lexicon is significantly more negatively valenced ($M = 0.44$) than general time expressions ($M = 0.23$) ($t = 12.54, p < .001$). Notably, it also showed a higher positive sentiment ($M = 0.26$) than the general time expressions ($M = 0.17$) ($t = 6.34, p < .001$). Conversely, general time expressions were significantly more neutral ($M = 0.60$) than the temporal lexicon ($M = 0.29$) ($t = -15.27, p < .001$). These results indicate that the language used to describe lived temporal experience in ASC is characterized by heightened emotional intensity, both negative and positive, with a particularly pronounced skew toward negative affect, whereas language used to describe time in a general sense (e.g., chronology of events) is predominantly neutral.

\subsection*{Study C: Methods}

\textbf{Sequentiality of Autistic Narratives.} \textcite{sap2022} introduced \textit{sequentiality}, a computational measure of narrative flow. Developed to compare how preceding sentences and the overall story topic influence a recalled versus imagined narrative, the measure relies on log probabilities extracted from a language model (originally GPT-3 \parencite{gpt3}). The study found that sequentiality values are higher in imagined stories than in autobiographical ones, suggesting that fictional narratives rely more on prior context and flow more predictably. To contextualize our previous findings on the narrative level of representation, we replicated a part of Sap et al.'s study on the ASC Corpus. The goal was to investigate whether autistic narratives, due to their specific content and writing style, are more or less grounded in the narrator's personal experiences than expected. We reconstructed the experimental environment and protocol, also using GPT-3 as the source of log probabilities, and similarly varied the input history size (measured in story sentences) from 1 to full. For comparison, we targeted the \textit{recalled} story category from the original study---denoting direct descriptions of autobiographical experiences. We treated the first 18 sentences of the main textual contents of a single book from our corpus as a story, thus aligning their length with the average 17.62 sentence length of the \textit{recalled} category stories.

\subsection*{Study C: Results}
\begin{figure}[htbp]
    \centering
    \includegraphics[width=\columnwidth, keepaspectratio]{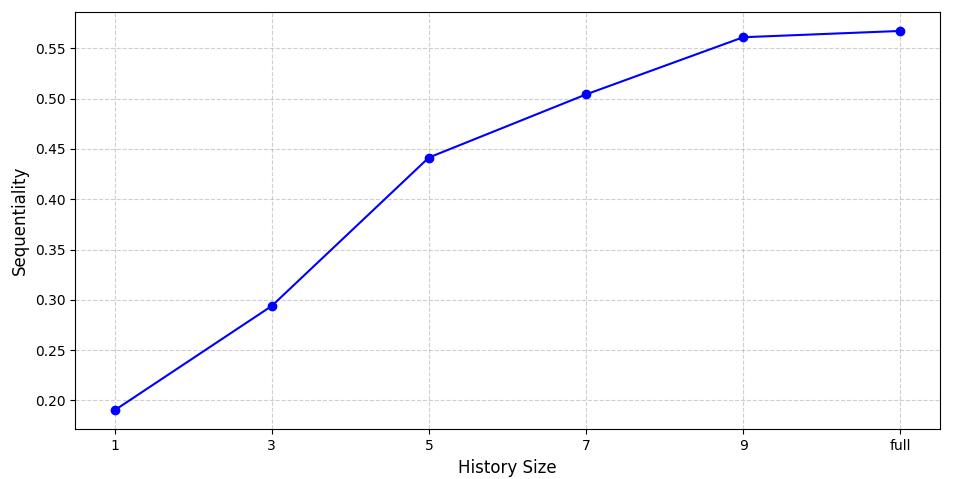}
    \caption{Average sequentiality by history size for the ASC Autobiographical Corpus.}
    \label{fig:sequentiality}
\end{figure}

\textbf{Sequentiality of Autistic Narratives. }Our replication on the ASC Corpus show that the experiences described therein resemble real and not imagined stories in terms of the averaged sequentiality value. Plateauing at 0.57, it is close to the value of above 0.6 for the recalled story category, as achieved by Sap et al. (as seen on a figure in the original study, with the precise value not given in the text). Therefore, it can be concluded that no specific ASC-oriented bias was identified for this metric (see Figure \ref{fig:sequentiality}).

\section{Discussion}
\textbf{Computational phenomenology.} Temporal continuity and predictability of events are crucial for the process of meaning-making and understanding the world, intentions, and behaviors of other subjects. Previous research clearly shows that both are disturbed in autism. However, there is a scarcity of large-scale phenomenological research to validate this from a first-person perspective using quantifiable measures. Our study exemplifies a practical application of computational phenomenology of temporal experience by sequentially combining a structured quantitative phenomenological interview (TATE) with corpus-based research on emotional intensity of temporal expressions and the narrative flow of autobiographical narratives in autism.  

\textbf{Front-loaded phenomenology of temporal expressions.} Although ,,representing time in language (...) is riddled with unresolved questions'' \parencite{jaszczolt2023}, our approach assumes that one may operationalize the findings of phenomenological psychopathology (TATE) for computational linguistics via a temporal lexicon (80 adverbs), and by analyzing the sentiments of the latter in a large-scale data source (ASC Autobiographical Corpus), as well as retrospectively interpret the phenomenon (lived experience of unpredictability) that was conceptualized philosophically with newly acquired empirical quantitative constraints. Following this approach, we operationalized two temporal phenomena of TATE, which refer to the past entering the present with immediacy (5.f) and sudden discontinuities of presence (6.k), with 80 temporal adverbs whose occurrences were identified in a corpus close to 7 million words in size. We found that negative sentiment prevailed across the temporal categories, while the category Sequence \& Relative Time proved to be the most prominent in terms of total mentions. The results indicated that first-person expressions of temporality in autism bear a heightened emotional intensity and are significantly skewed toward the negative affect. Simultaneously, the descriptions of time in a chronological sense are predominantly neutral. 
{\raggedbottom

\textbf{Suddenness of autistic experience.} The most important finding, however, is that both groups of observed outliers (positive and negative) refer to ,,sudden'' temporal states. That is, they indicate that experiences of transition (a kind of ,,transgression'', rather than a state of prolonged experience of a given situation) are strongly emotionally charged. This indicates that it is not the pace of the passage of time (typically referred to as \textit{flow}) that is important in autism, but the structure of transitions between moments (sudden, jumpy, fragmented). 

In this respect, it is worth noting that in the analysis of temporal categories, Pace stood as the only exception to the otherwise predominant negative sentiments (where the category of  Immediacy \& Suddenness had the highest average negative sentiment score), displaying a more balanced sentiment profile. What determines whether a given experience is classified as positive or negative? The ,,suddenness'' points to what phenomenology terms ,,prereflective'' experiences. As some research on the experience of time in autism indicates, if something happens \textit{swiftly} or \textit{instantly} (positive terms), it means it occurs according to intention, the accepted assumption, or a plan. The subject of these events has a sense of agency and understands the meaning of the events that come about smoothly. On the other hand, events that happen \textit{precipitously}, \textit{abruptly}, \textit{unpredictably} (negative categories in our sample) remain beyond control. The contents of these experiences are discontinuous. The subject feels ,,carried away'' by events and time. The happening does not depend on them; it merely happens to them. 

Both categories, positive and negative, indicate the singularity of an event (which is also noticeable in research on lived time experience in autism, which can be fragmented and ,,frayed'', disrupting the sense of security and the ability to understand the world around). Routines and rituals are designed to counteract this, restoring temporal continuity and providing a basis for predictability. Unfortunately, the world of interpersonal relationships cannot be easily confined to such routine and predictability.

\textbf{Genuine experience and not narrative construction.} Importantly, mutually reinforcing observations can be abstracted away from the computational results of Studies B and C, even though these are methodologically distinct. The fine-grained units of valence quantified through the sentiment values of the temporal expressions in short narrative units (sentence triplets) can be now interpreted in light of the values of the dedicated measure (sequentiality) of longer narrative flow (stories up to 18 sentences long) achieved for fragments from the corpus in Study C. Seen together, the results of both studies support the argument that temporal challenges of autistic individuals stem from the contents of their lived experience and not mere narrative construction. In other words, the discrete (as seen from the point of view of the entire narrative flow) negative experiences of ,,suddenness'' are only singular points of the narrative and do not disturb its overall coherence to a level at which it would noticeably (computationally or else) diverge from its neurotypical equivalents, thus seeming inauthentic or imagined. 
}

% Conclusion - BY GEMINI
%\section{Conclusion}
%This research integrated structured phenomenological inquiry with large-scale computational linguistics to provide a nuanced, data-driven account of how time is experienced in autism. Our findings indicate that temporal challenges in ASC are primarily driven by \textbf{lived unpredictability}—a fragmented experience of transitions rather than a simple distortion of time's pace. 
%
%Sentiment analysis of the ASC Autobiographical Corpus reveals that temporal expressions are significantly more negatively valenced than general time expressions, with outliers like ``unpredictably'' and ``abruptly'' highlighting a disrupted sense of agency. Crucially, the sequentiality analysis suggests that these challenges are grounded in the authentic contents of lived experience rather than a deficit in narrative construction. By scaling first-person accounts through computational phenomenology, this study moves beyond medical deficit models to validate the subjective reality of the neurodiverse experience.

\section{Conclusion}
This study reinforces the scientific validity of the claim that the primary experiential challenge in autism concerns the lived unpredictability of experience. It does so both phenomenologically, that is, via structured interviews with autistic individuals, and computationally, that is, by exploring a 7-million-word corpus of autistic autobiographies. The key results show that sudden transitory experiences carry a significantly higher negative valence for autistic individuals. By contrasting these findings with measures of narrative flow, the study demonstrates that the disturbing nature of these experiences is not a feature of their autobiographical storytelling, but a reflection of lived experience preserved in the semantics of temporal expressions. 

\section{Limitations}
The two main limitations of this research are specifically related to temporality. First, the TATE interview is cross-sectional and targets discrete temporal phenomena, but its conclusions are extrapolated to structures of consciousness (that is, treated as if they represented such structures). Second, ASC corpus analysis is conducted on retrospective narrative representations of lived experiences, and not on direct qualitative reports. The conclusions, however, are also extrapolated from this higher representational level to that of lived experience. In other words, both tools are retrospective and do not directly target the lived experience in its actual flow (unlike, e.g., microphenomenology).  

\printbibliography

% --- START OF APPENDICES ---
\clearpage
\onecolumn % Force single column layout for the appendices

\appendix

% ==========================================
% NEW APPENDIX A: Factor Analysis (Figures 1 & 2)
% ==========================================
\refstepcounter{section} % Increment to A
\section*{Appendix \thesection: Factor Analysis and Group Differences}
\addcontentsline{toc}{section}{Appendix \thesection: Factor Analysis and Group Differences}
\label{app:factor_analysis}

% Reset figure counter for Appendix A
\setcounter{figure}{0}
\renewcommand{\thefigure}{\thesection\arabic{figure}}

\begin{figure}[h!]
    \centering
    % First Figure (Standard width)
    \includegraphics[width=0.85\textwidth, height=0.4\textheight, keepaspectratio]{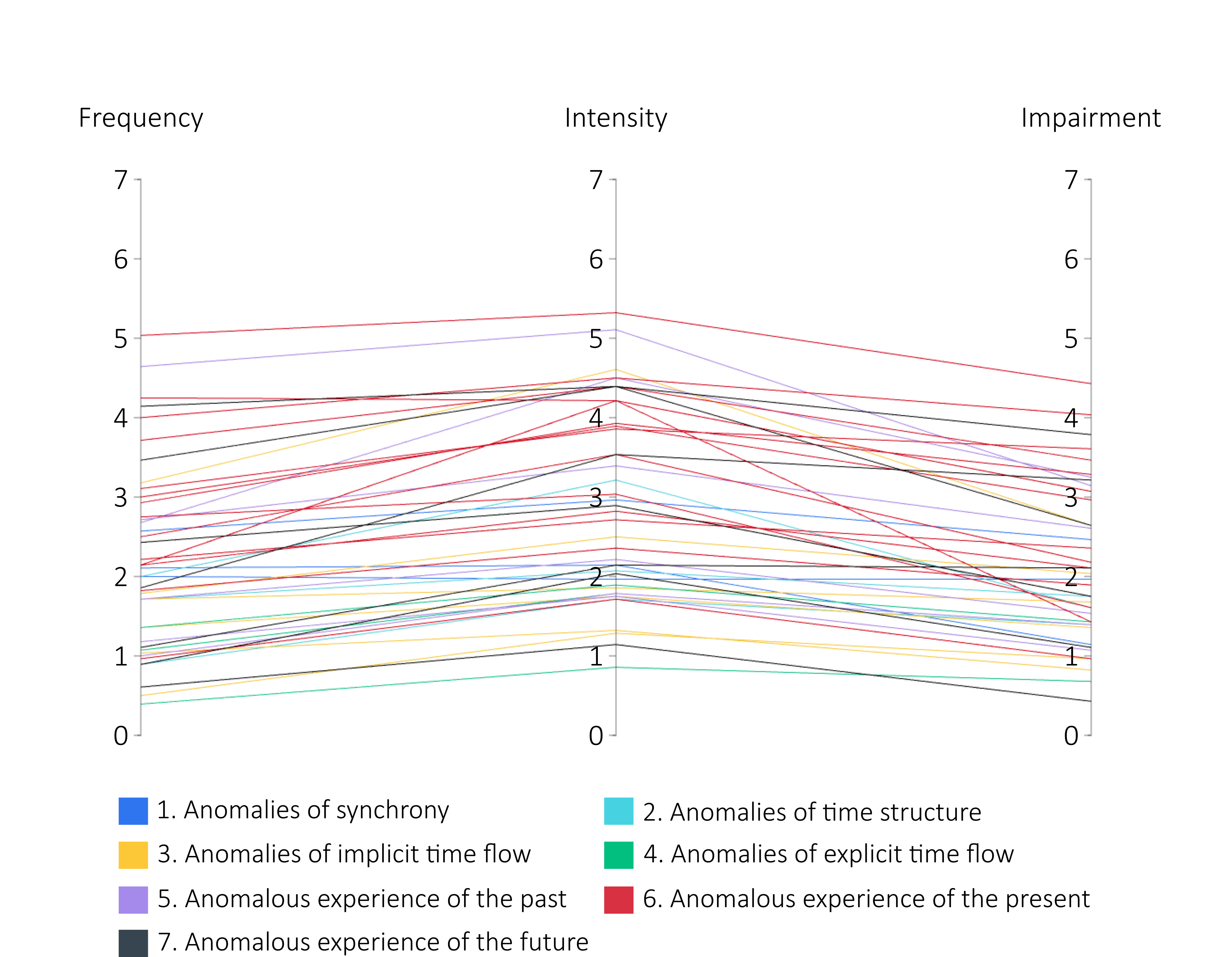}
    \caption{Seven dimensions of temporal experience as measured by TATE on three scales (Frequency, Intensity and Impairment).}
    \label{fig:factor_analysis}
\end{figure}

\vspace{1em}

\begin{figure}[h!]
    \centering
    % Second Figure (Reduced width to appear visually similar to the first)
    \includegraphics[width=0.65\textwidth, height=0.4\textheight, keepaspectratio]{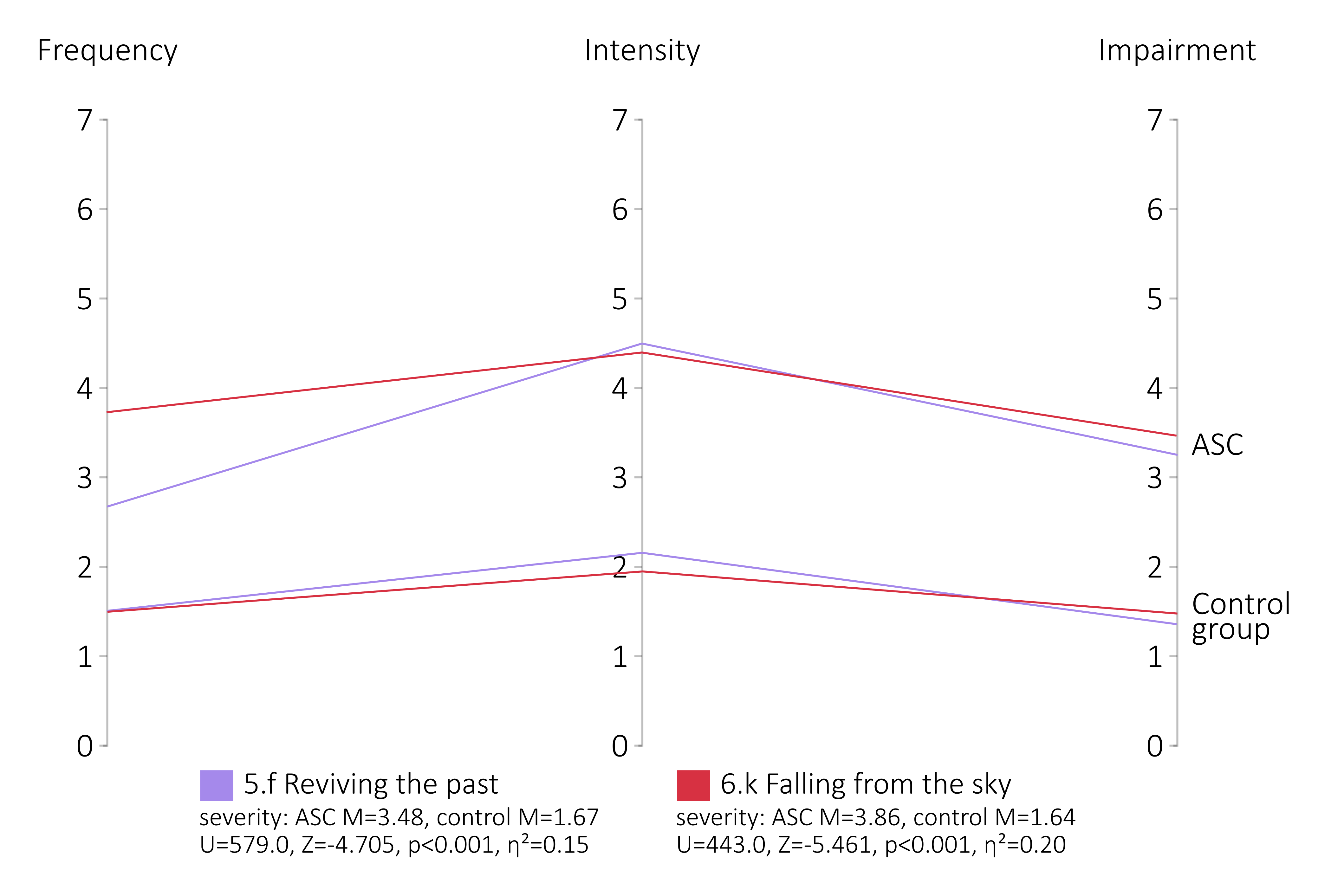}
    \caption{Group differences in TATE item scores (5.f. and 6.k.) between ASC and control group.}
    \label{fig:group_differences}
\end{figure}

% ==========================================
% APPENDIX B: TATE Items (Formerly A)
% ==========================================
\clearpage
\refstepcounter{section} % Increment to B
\section*{Appendix \thesection: Transdiagnostic Assessment of Temporal Experience (TATE) Items}
\addcontentsline{toc}{section}{Appendix \thesection: Transdiagnostic Assessment of Temporal Experience (TATE) Items}
\label{app:tate_items}

% Reset table counter for Appendix B
\setcounter{table}{0}
\renewcommand{\thetable}{\thesection\arabic{table}}

\begin{table}[h!]
    \centering
    \renewcommand{\arraystretch}{1.5}
    \setlength{\tabcolsep}{12pt}
    \begin{tabular}{@{}p{0.95\textwidth}@{}}
        \toprule
        \textbf{Item 5.f: Reviving the Past} \\
        \midrule
        \textit{Do you ever experience sudden memories with such intensity that it seems as if the past were happening now?} \\
        \vspace{0.5em}
        \textbf{Guiding Questions:} Think about whether\dots \\
        \begin{itemize}[noitemsep, topsep=0pt, leftmargin=1.5em]
            \item \dots you have sudden, intense memories that you experience strongly?
            \item \dots you experience past events with the same force as when they happened?
        \end{itemize} \\
        \bottomrule
    \end{tabular}
\end{table}

\vspace{2em}

\begin{table}[h!]
    \centering
    \renewcommand{\arraystretch}{1.5}
    \setlength{\tabcolsep}{12pt}
    \begin{tabular}{@{}p{0.95\textwidth}@{}}
        \toprule
        \textbf{Item 6.k: Falling from the Sky} \\
        \midrule
        \textit{Do you ever get so absorbed in yourself and your thoughts that you are surprised by what is happening around you?} \\
        \vspace{0.5em}
        \textbf{Guiding Questions:} Think about whether\dots \\
        \begin{itemize}[noitemsep, topsep=0pt, leftmargin=1.5em]
            \item \dots you sometimes feel that everything that happens is a surprise to you?
            \item \dots you ever get surprised by what happens to you as something unexpected?
        \end{itemize} \\
        \bottomrule
    \end{tabular}
\end{table}

% ==========================================
% APPENDIX C: Lexicon (Formerly B)
% ==========================================
\clearpage
\refstepcounter{section} % Increment to C
\section*{Appendix \thesection: Lexicon of Time Adverbs}
\addcontentsline{toc}{section}{Appendix \thesection: Lexicon of Time Adverbs}
\label{app:lexicon}

\noindent
This appendix presents a comprehensive inventory of time adverbs (N = 80) selected to capture the experience of time and temporal processing, organized into six semantic categories.

\vspace{1em}

\begin{table}[h!]
    \centering
    \renewcommand{\arraystretch}{1.3}
    \small
    \begin{tabularx}{\textwidth}{@{} l X @{}}
        \toprule
        \textbf{Category} & \textbf{Adverbs} \\
        \midrule
        
        \textbf{Immediacy \& Suddenness} \newline (N = 13) & 
        Abruptly, Directly, Immediately, Instantly, Instantaneously, Precipitously, Promptly, Spontaneously, Straightaway, Suddenly, Unexpectedly, Unforeseeably, Unpredictably \\
        \addlinespace[0.5em]
        
        \textbf{Frequency \& Repetition} \newline (N = 21) & 
        Always, Chronically, Constantly, Cyclically, Ever, Frequently, Generally, Intermittently, Never, Normally, Occasionally, Often, Periodically, Rarely, Recurrently, Regularly, Repeatedly, Sometimes, Sporadically, Usually, Whenever \\
        \addlinespace[0.5em]
        
        \textbf{Duration \& Time Blindness} \newline (N = 16) & 
        Briefly, Continually, Continuously, Endlessly, Fleetingly, Forever, Indefinitely, Interminably, Momentarily, Overnight, Permanently, Perpetually, Shortly, Still, Temporarily, Transiently \\
        \addlinespace[0.5em]
        
        \textbf{Sequence \& Relative Time} \newline (N = 23) & 
        \textit{Past:} Ago, Before, Last, Lately, Previously, Recently, Yesterday, Retrospectively, Since \newline
        \textit{Present:} Currently, Now, Presently, Today \newline
        \textit{Future/Sequence:} After, Already, Eventually, Lastly, Later, Next, Soon, Then, Yet, Belatedly \\
        \addlinespace[0.5em]
        
        \textbf{Pace} \newline (N = 5) & 
        Gradually, Quickly, Rapidly, Slowly, Swiftly \\
        \addlinespace[0.5em]
        
        \textbf{Overlap} \newline (N = 2) & 
        Concurrently, Simultaneously \\
        
        \bottomrule
    \end{tabularx}
\end{table}

% ==========================================
% APPENDIX D: Sentiment Distribution (Formerly C)
% ==========================================
\clearpage
\refstepcounter{section} % Increment to D
\section*{Appendix \thesection: Sentiment Distribution}
\addcontentsline{toc}{section}{Appendix \thesection: Sentiment Distribution}
\label{app:sentiment_dist}

% Reset figure counter for Appendix D
\setcounter{figure}{0}
\renewcommand{\thefigure}{\thesection\arabic{figure}}

\begin{figure}[h!]
    \centering
    \includegraphics[width=0.85\textwidth]{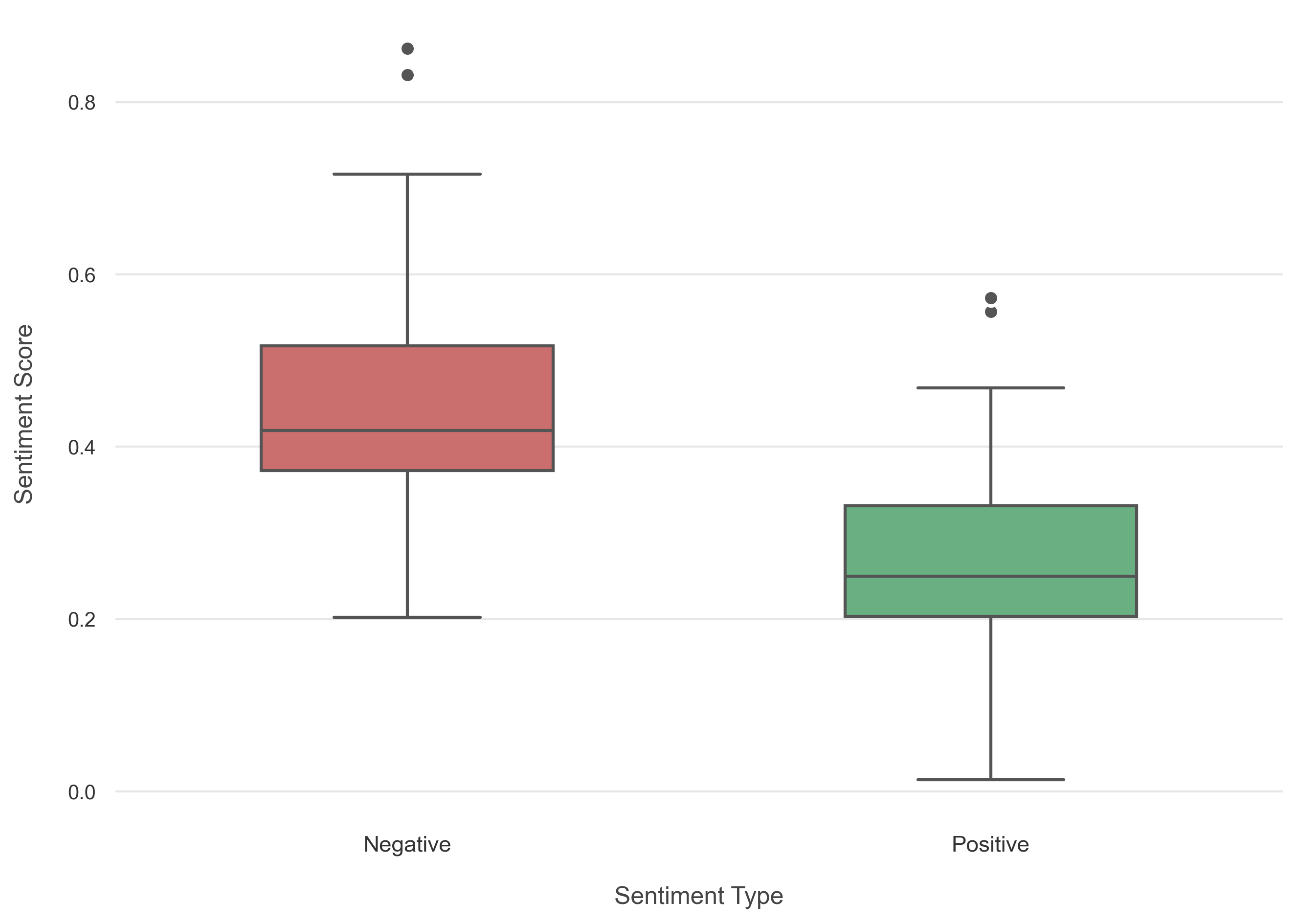}
    \caption{Distribution difference between positive and negative sentiments.}
    \label{fig:sentiment_boxplot}
\end{figure}

% ==========================================
% APPENDIX E: Contextual Examples (Formerly D)
% ==========================================
\clearpage
\refstepcounter{section} % Increment to E
\section*{Appendix \thesection: Contextual Examples of Sentiment Outliers}
\addcontentsline{toc}{section}{Appendix \thesection: Contextual Examples of Sentiment Outliers}
\label{app:examples}

% Reset table counter for Appendix E
\setcounter{table}{0}
\renewcommand{\thetable}{\thesection\arabic{table}}

\begin{table}[h!]
    \centering
    \renewcommand{\arraystretch}{1.5}
    \small
    \begin{tabularx}{\textwidth}{@{} l X @{}}
        \toprule
        \textbf{Word} & \textbf{Contextual Excerpt from ASC Corpus} \\
        \midrule
        
        \multicolumn{2}{@{}l}{\textit{\textbf{Negative Sentiment Outliers}}} \\
        \addlinespace[0.5em]
        
        \textbf{Precipitously} & 
        ``These flattering gestures planted the seeds of confidence and optimism for the future. My quasi-good looks led to plenty of infatuations, which were \textbf{precipitously} halted once my female suitor detected something was a little . . . off. Her phone calls and emails would end, as she always sent the message by `ignoring me'.'' \newline \textit{(Source ID: 62178)} \\
        \addlinespace[1em]

        \textbf{Unpredictably} & 
        ``I struggle with clothing tags, which are made of different material to the clothing they are fitted into, and stick out at odd angles, flapping around \textbf{unpredictably} and poking into me, dragging uncomfortably and taking up all my attention trying to manage.'' \newline \textit{(Source ID: 104520)} \\
        \addlinespace[1em]

        \textbf{Abruptly} & 
        ``I suppose I had given some involuntary signal that I was fixating on certain letters and combinations, probably because the stroke of my finger changed. When the Canon was turned over, the keys \textbf{abruptly} vanished, and the sudden tunnel through which I had started to make a swooping, homing motion decomposed.'' \newline \textit{(Source ID: 31532)} \\
        
        \midrule
        \multicolumn{2}{@{}l}{\textit{\textbf{Positive Sentiment Outliers}}} \\
        \addlinespace[0.5em]

        \textbf{Swiftly} & 
        ``After a few minutes, a blond-haired boy entered the room and apologized for his being late. He scanned the classroom \textbf{swiftly} and spotted the empty chair next to me. I was immediately relieved.'' \newline \textit{(Source ID: 17119)} \\
        \addlinespace[1em]

        \textbf{Instantaneously} & 
        ``But I coped OK. It was whilst I was there that I made the discovery that food can make me \textbf{instantaneously} happy. I’d like to go to Japan now; the culture really appeals to me.'' \newline \textit{(Source ID: 47060)} \\
        \addlinespace[1em]

        \textbf{Instantly} & 
        ``By making this decision I felt a huge shift in responsibility, from the maternity service to myself and this felt very empowering. I \textbf{instantly} dropped my previous interests and began to devour all the information I could dig up on pregnancy and birth...'' \newline \textit{(Source ID: 134101)} \\

        \bottomrule
    \end{tabularx}
    \caption{Contextual examples of high-value sentiment outliers.}
    \label{tab:examples}
\end{table}

% ==========================================
% APPENDIX F: Lexical Diversity (Formerly E)
% ==========================================
\clearpage
\refstepcounter{section} % Increment to F
\section*{Appendix \thesection: Lexical Diversity and Frequency}
\addcontentsline{toc}{section}{Appendix \thesection: Lexical Diversity and Frequency}
\label{app:frequency}

% Reset figure counter for Appendix F
\setcounter{figure}{0}
\renewcommand{\thefigure}{\thesection\arabic{figure}}

\begin{figure}[h!]
    \centering
    \includegraphics[width=0.9\textwidth]{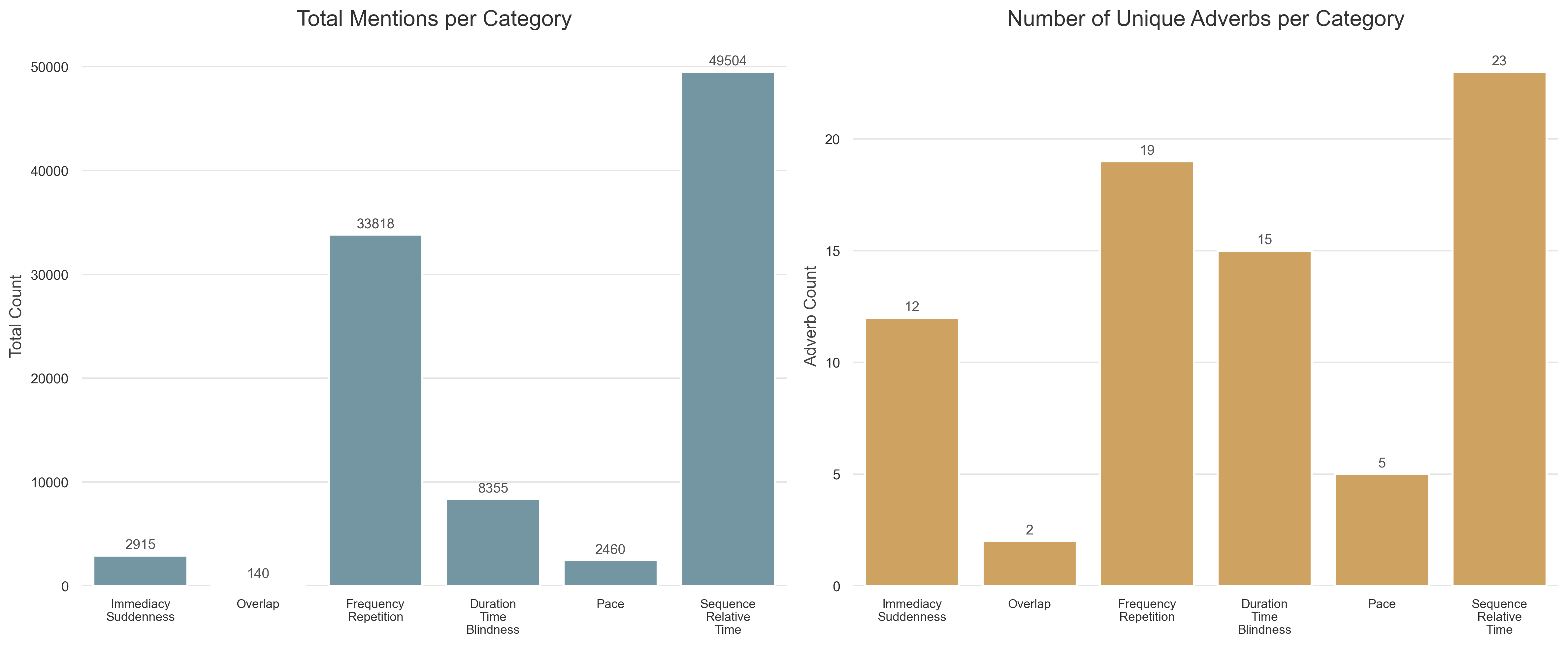}
    \caption{Lexical diversity and usage frequency across the six temporal categories.}
    \label{fig:frequency}
\end{figure}

% --- END OF APPENDICES ---

\end{document}